\pdfoutput=1

\documentclass[11pt]{article}

\usepackage[]{EMNLP2022}

\usepackage{times}
\usepackage{latexsym}

\usepackage[T1]{fontenc}

\usepackage[utf8]{inputenc}

\usepackage{microtype}

\usepackage{inconsolata}

\usepackage{graphicx}
\usepackage{booktabs,caption}
\usepackage{multirow}
\usepackage{enumitem}
\usepackage{textcomp}
\usepackage{makecell}
\usepackage{arydshln}
\usepackage{amsmath}
\usepackage{stfloats}
\usepackage{soul}
\usepackage[capitalise]{cleveref}
\usepackage{hyperref}

\DeclareMathOperator*{\argmin}{arg\,min}
\DeclareMathOperator{\myst}{s.t.}
\definecolor{hzw}{RGB}{223, 97, 76}

\newcommand{\footlabel}[2]{%
    \addtocounter{footnote}{1}%
    \footnotetext[\thefootnote]{%
        \addtocounter{footnote}{-1}%
        \refstepcounter{footnote}\label{#1}%
        #2%
    }%
    $^{\ref{#1}}$%
}

\newcommand{\myfootref}[1]{%
    $^{\ref{#1}}$%
}

\title{Tencent AI Lab - Shanghai Jiao Tong University Low-Resource Translation System for the WMT22 Translation Task}

\author{
Zhiwei He\thanks{\ \ Work was done when Zhiwei He was interning at Tencent AI Lab.}\\Shanghai Jiao Tong University\\ \normalsize \sf zwhe.cs@sjtu.edu.cn \And
Xing Wang\thanks{\ \ Xing Wang is the corresponding author.}\\Tencent AI Lab\\ \normalsize \sf brightxwang@tencent.com \AND
Zhaopeng Tu\\Tencent AI Lab\\ \normalsize \sf zptu@tencent.com \And
Shuming Shi\\Tencent AI Lab\\ \normalsize \sf shumingshi@tencent.com \And
Rui Wang\\Shanghai Jiao Tong University\\ \normalsize \sf wangrui12@sjtu.edu.cn
}

\begin{document}
\maketitle
\begin{abstract}
This paper describes Tencent AI Lab - Shanghai Jiao Tong University (TAL-SJTU) Low-Resource Translation systems for the WMT22 shared task. 
We participate in the general translation task on English$\Leftrightarrow$Livonian.
Our system is based on M2M100~\cite{JMLR:v22:20-1307} with novel techniques that adapt it to the target language pair.
(1) \ul{Cross-model word embedding alignment}: inspired by cross-lingual word embedding alignment, we successfully transfer a pre-trained word embedding to M2M100, enabling it to support Livonian.
(2) \ul{Gradual adaptation strategy}: we exploit Estonian and Latvian as auxiliary languages for many-to-many translation training and then adapt to English-Livonian.
(3) \ul{Data augmentation}: to enlarge the parallel data for English-Livonian, we construct pseudo-parallel data with Estonian and Latvian as pivot languages.
(4) \ul{Fine-tuning}: to make the most of all available data, we fine-tune the model with the validation set and online back-translation, further boosting the performance.
In model evaluation: (1) We find that previous work~\cite{rikters-etal-2022-machine} underestimated the translation performance of Livonian due to \ul{inconsistent Unicode normalization}, which may cause a discrepancy of up to 14.9 BLEU score.
(2) In addition to the standard validation set, we also employ \ul{round-trip BLEU} to evaluate the models, which we find more appropriate for this task. 
Finally, our unconstrained system achieves BLEU scores of 17.0 and 30.4 for English to/from Livonian.\footnote{\ Code, data, and trained models are available at \url{https://github.com/zwhe99/WMT22-En-Liv}.}

\end{abstract}

\section{Introduction}
This paper introduces our submissions to the WMT22 general machine translation task.
Last year, Tencent AI Lab participated in two translation tasks: News~\cite{wang2021tencent1} and Biomedical translation~\cite{wang2021tencent2}.
This year, we participate in English$\Leftrightarrow$Livonian (En$\Leftrightarrow$Liv), a very low-resource and distant language pair.
Considering the scarcity of parallel En-Liv corpus, we only participate in the unconstrained evaluation.

We use M2M100 1.2B\footnote{\url{https://github.com/facebookresearch/fairseq/tree/main/examples/m2m_100}}~\cite{JMLR:v22:20-1307} as the pre-trained model which is a massive multilingual translation model that supports any pair of 100 languages\footnote{M2M100 supports English, Latvian and Estonian.} and shows promising performance for low-resource translation.
To adapt it to En-Liv, the first thing to do is enabling it to support Liv.
A common approach is to expand the vocabulary and the word embedding matrix to contain the extra tokens.
However, the incoming embeddings must be randomly initialized~\cite{garcia-etal-2021-towards,bapna2022building}, which leads to inconsistency with the original embeddings and increases training difficulty.
Fortunately, \citet{rikters-etal-2022-machine} has released a translation model for En-Liv called Liv4ever-MT\footlabel{note:liv4ever-mt}{\url{https://huggingface.co/tartuNLP/liv4ever-mt}}.
Inspired by supervised cross-lingual word embedding alignment~\cite{lample2018word}, we propose cross-model word embedding alignment (CMEA) that learns a linear transformation between the embedding matrices of two models.
Therefore, the incoming embeddings can be extracted from Liv4ever-MT and transformed to M2M100's word embedding space rather than random initialization.

In terms of model training, we adopt a gradual adaptation strategy.
The overall training process is shown in \figurename~\ref{fig:training-process}.
Following~\citet{rikters-etal-2022-machine}, we also use Estonian (Et) and Latvian (Lv) as auxiliary languages.
Liv has been influenced by Et and Lv for centuries.
There are about 800 Et loanwords and 2,000 Lv loanwords in Liv~\cite{decsy1965einfuhrung}.
Therefore, we first add Et and Lv
for many-to-many translation training, resulting in a 4-lingual translation model.
We then augment the En-Liv data with forward and backward translations using Et and Lv as the pivot languages.
Finally, we combine all the authentic and synthetic data to retrain the model, followed by a few steps of fine-tuning with the validation set and online back-translation.

In terms of model evaluation, we find that the data set provided by~\citet{rikters-etal-2022-machine} suffers from inconsistent Unicode normalization.
This inconsistency is reflected in using two or more encodings for the same character, which leads to inconsistent encoding between model hypothesis and reference\footnote{SentencePiece does uniform normalization by default. Therefore, the character encoding in the model hypothesis is uniform but may not be consistent with the reference.} and thus inaccurate evaluation.
In our experiments, normalizing the character encoding can bring an average improvement of +2.5 BLEU on the liv4ever\footlabel{note:liv4ever}{\url{https://opus.nlpl.eu/liv4ever-v1.php}} test set (see appendix \ref{sec-appendix:re-evluating-liv4ever-mt}) and up to +14.9 BLEU on a subset from a specific source.
In addition to the standard validation set, we also employ round-trip BLEU to evaluate our models, which is an effective unsupervised criterion~\cite{lample2018unsupervised} and reduces the demand for the parallel corpus.
\citet{zhuo2022rethinking} have found that
in the scope of neural machine translation, round-trip translation quality correlates consistently with forward translation quality.
We consider round-trip BLEU a better evaluation method for this task.
The reasons for this are threefold: more data, more general domain, and the same original language as the WMT22 En-Liv test set.

This paper is structured as follows:
\cref{sec:data-and-processing} describes the data statistics and processing methods.
Then we present our evaluation methods in \cref{sec:model-evaluation}.
Our translation system and ablation study are detailed in \cref{sec:system-and-ablation-study}, followed by the final results.
Finally, we conclude the paper in \cref{sec:conclusion}.

\section{Data and Processing}
\label{sec:data-and-processing}
\subsection{Overview}

\paragraph{Statistics}
\tablename~\ref{tab:data-statistics} lists statistics of the parallel and monolingual data we used. We collect parallel data for any pair in $\{$En, Liv, Et, Lv$\}$ and collect monolingual data for En and Liv.

\begin{table}[htpb]  
    \centering
    \setlength\tabcolsep{3.3pt}
    \begin{tabular}{c l r r}
        \toprule
       \multirow{2}{*}{\bf Data } & \multirow{2}{*}{\bf Lang} & \multicolumn{2}{c}{\bf \# Sent.}\\
        \cmidrule(lr){3-4} 
        & & \bf Raw & \bf Filter  \\
        \midrule
        \multirow{6}{*}{Parallel Data} & En-Liv & 1.2K  & 1.1K \\ 
         & En-Et  & 40.3M & 20.7M \\ 
         & En-Lv  & 27.2M & 11.3M \\
         & Liv-Et & 14.8K & 14.8K \\
         & Liv-Lv & 12.4K & 12.2K \\
         & Et-Lv  & 10.7M  & 7.0M \\
         \midrule
          \multirow{2}{*}{Monolingual Data} & En & 325.6M & 281.3M\\
          & Liv & 138.2K & 50.2K \\
        \bottomrule
    \end{tabular}
    \caption{Statistics of parallel and monolingual data. We report the number of sentences before and after filtering.}
    \label{tab:data-statistics}
\end{table}

\paragraph{Data Source}
The parallel data is mainly all available corpora from OPUS\footnote{\url{https://opus.nlpl.eu/}}.
Due to the scarcity of data, we include liv4ever-dev in training data and use liv4ever-test as the validation set.
For En-Et and En-Lv, we augment them with the parallel data from WMT18 and WMT17, respectively.
For En-Liv, En-Lv and Liv-Lv, we collected additional parallel data from Facebook posts of the Livonian Institute and Livones.net\footnote{The numbers of additional sentences collected from Facebook are En-Liv: 54, En-Lv: 61 and Liv-Lv: 61.}.
The monolingual En is News Crawl 2007-2021.
The monolingual Liv combines all Liv from parallel data and monolingual data from liv4ever\myfootref{note:liv4ever}.

\subsection{Pre-processing}
To obtain higher quality training data, we employ a series of data cleaning using Moses toolkit\footnote{\url{https://github.com/moses-smt/mosesdecoder}} and our scripts\footnote{\url{https://github.com/zwhe99/corpus-tools}}.
We process parallel data as follows:
\begin{itemize}[leftmargin=20pt,parsep=5pt,itemsep=0pt,topsep=5pt]
    \item Replace Unicode punctuation, normalize punctuation and remove non-printing characters
    \item Language identification and filtering
    \item Remove instances with too much punctuation
    \item Remove instances with identical source and target sentences
    \item Remove instances containing URLs
    \item Remove instances appearing in evaluation data
    \item Remove instances with more than 175 tokens or length ratio over 1.5
\end{itemize}
The liv4ever corpus has a small amount of data, and the existing tools may not support Liv well.
Therefore, for the liv4ever corpus, we don't apply punctuation processing or language and length ratio filtering.
For the monolingual data, we use the same cleaning steps as parallel data except for identical source-target filtering and length ratio filtering.

After cleaning the data, we apply SentencePiece\footnote{\url{https://github.com/google/sentencepiece}} encoding using the trained model from Liv4ever-MT\myfootref{note:liv4ever-mt}.
We also reuse their vocabulary that shared by all languages.

\subsection{Evaluation Data}
We regard the liv4ever-test as the validation set, which is a multi-way data set for $\{$En, Liv, Et, Lv$\}$ containing 855 unique sentences.
Besides, for En$\Leftrightarrow$Liv evaluation,  we collect monolingual English from the source of WMT22 English-German (En-De) test set to compute round-trip BLEU (En$\Rightarrow$Liv$\Rightarrow$En).

\section{Model Evaluation}
\label{sec:model-evaluation}
This section describes our methods for model evaluation.
Specifically, we explain the Unicode inconsistency problem in the liv4ever data set and the resulting underestimation of model performance.
In addition, we introduce round-trip BLEU as the more appropriate way for this competition.

\subsection{Unicode inconsistency problem}
\citet{rikters-etal-2022-machine} collected the liv4ever data set and built Liv4ever-MT, the first machine translation model for Livonian.
We find that the liv4ever data set does not use consistent Unicode normalization, resulting in different encodings for the same character.
This did not lead to any training problem in \citet{rikters-etal-2022-machine} because SentencePiece does NFKC\footnote{\url{https://unicode.org/reports/tr15/}} normalization by default.
However, when computing SacreBLEU\footnote{\url{https://github.com/mjpost/sacrebleu}}, the encoding of model output and the reference will be inconsistent, resulting in inaccurate evaluation.

We re-evaluate the performance of Liv4ever-MT before and after normalizing the encoding of references to NFKC.
\tablename~\ref{tab:unicode-nfkc} shows the SacreBLEU results\footnote{nrefs:1|case:mixed|eff:no|tok:13a|smooth:exp|version:2.0.0} on the entire test set and a subset from Satversme.
Before normalization, our results are very close to those reported in \citet{rikters-etal-2022-machine}, while after normalization, the BLEU score improves considerably.
In particular, the difference in BLEU score is up to 14.9 on the Lv$\Rightarrow$Liv of the Satversme subset.
Therefore, we report SacreBLEU after normalization in the following.

\begin{table}[htpb]  
    \centering
    \setlength\tabcolsep{3pt}
    \begin{tabular}{l c c c c c c }
        \toprule
         & \multicolumn{2}{c}{\bf En-Liv} & \multicolumn{2}{c}{\bf Et-Liv} & \multicolumn{2}{c}{\bf Lv-Liv} \\
        \cmidrule(lr){2-3} \cmidrule(lr){4-5} \cmidrule(lr){6-7} 
        & $\Rightarrow$ & $\Leftarrow$ & $\Rightarrow$ & $\Leftarrow$ & $\Rightarrow$ & $\Leftarrow$ \\
        \midrule
        \multicolumn{7}{c}{\bf All} \\
        \makecell[l]{\bf Liv4ever-MT\\(\citeauthor{rikters-etal-2022-machine})} &  11.0 & 19.0 & 16.5 & 23.1 & 17.7 & 25.2\\
        \hdashline
        \bf Our Eval.     &      10.9 & 18.9 & 16.6 & 22.9 & 17.7 & 24.9 \\
        \bf + Norm. Ref.  &  \bf14.3 & \bf19.3 & \bf20.5 & \bf24.4 & \bf22.3 & \bf29.3 \\

        \midrule
        \multicolumn{7}{c}{\bf Subset (Satversme)} \\
        \makecell[l]{\bf Liv4ever-MT\\(\citeauthor{rikters-etal-2022-machine})} &  7.7 & 24.5 & - & - & - & - \\
        \hdashline
        \bf Our Eval.     & 7.6  & 24.7 & 7.2  & 18.7 & 9.2  & 19.4      \\
        \bf + Norm. Ref.  & \bf18.2 & \bf25.8 & \bf19.9 & \bf23.7 & \bf24.2 & \bf33.6      \\
        \bottomrule

    \end{tabular}
    \caption{BLEU scores of Liv4ever-MT on liv4ever-test. \textbf{Liv4ever-MT}~(\citeauthor{rikters-etal-2022-machine}): copied from~\citet{rikters-etal-2022-machine}. \textbf{Our Eval.}: We use the released Liv4ever-MT to generate translation outputs and re-evaluate them with the original references, which shows similar results compared with \citet{rikters-etal-2022-machine}.
    \textbf{+ Norm. Ref.}: re-evaluation after normalizing the encoding of references to NFKC.
    See \cref{sec-appendix:re-evluating-liv4ever-mt} for all language pairs.}
    \label{tab:unicode-nfkc}
\end{table}

\subsection{Round-trip BLEU}
\label{sec:round-trip-bleu}
We collect monolingual English from the source of WMT22 English-German (En-De) test set and conduct two steps translation: En$\Rightarrow$Liv$\Rightarrow$En.
The round-trip BLEU score can be obtained by comparing the original input with the model output English.
We regard it a better way to evaluate En$\Leftrightarrow$Liv performance for this task considering three aspects: 
(1) En-De test set has 20683 sentences, much more than the liv4ever-test.
(2) It may contain more general domain data, while the liv4ever-test is relatively restricted due to the low-resource limitation.
(3) The original language used in computing the round-trip BLEU is the same as the WMT22 En-Liv test set (both English-original).

\section{System and Ablation Study}
\label{sec:system-and-ablation-study}
In this section, we describe our system in this competition and provide a comprehensive ablation study of the key components.

\subsection{System Overview}
\label{sec:system-overview}
\begin{figure*}[htpb]
    \centering
    \includegraphics[width=0.8\textwidth]{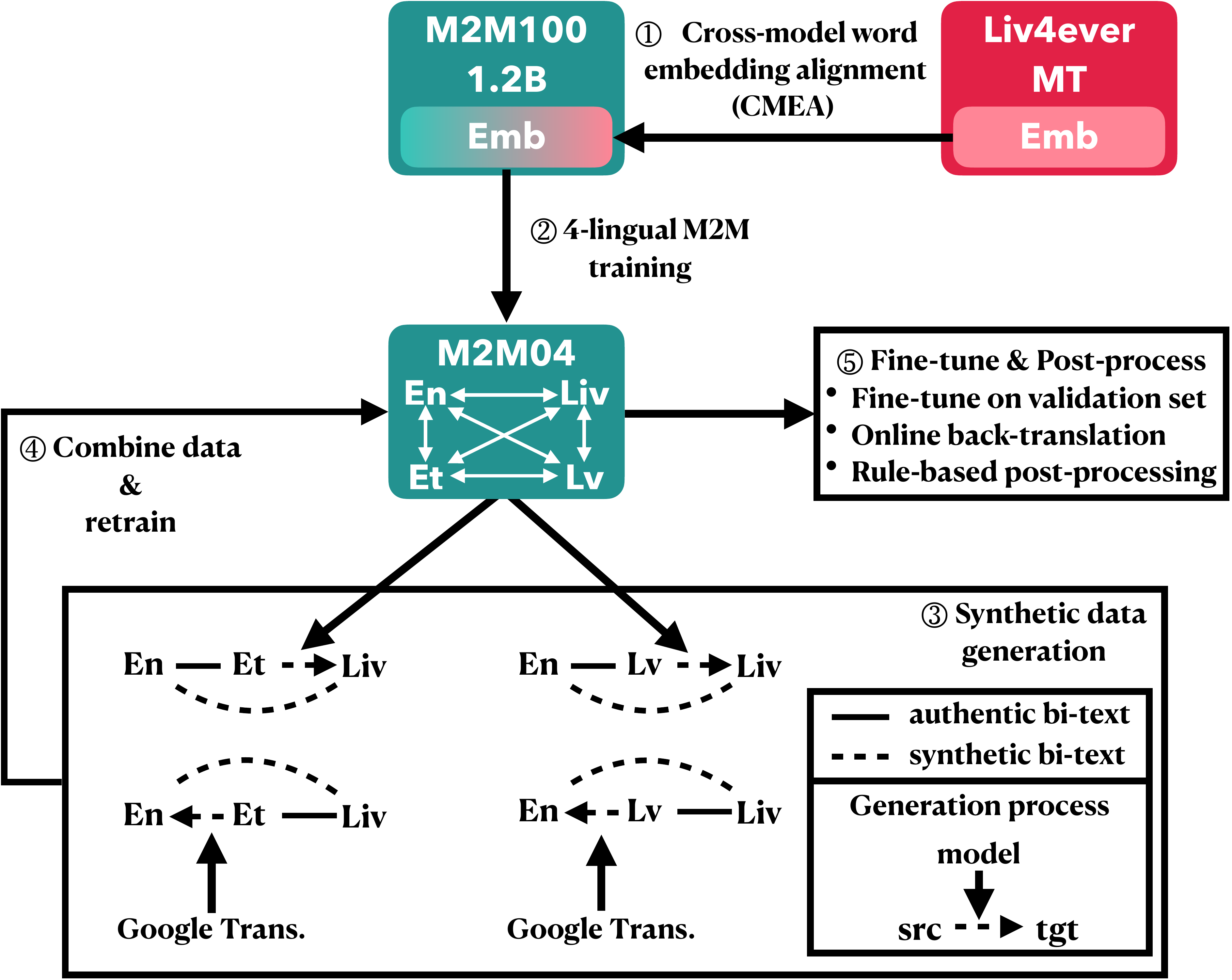}
    \caption{The training process of our translation system.}
    \label{fig:training-process}
\end{figure*}
We depict the overview of our system in \figurename~\ref{fig:training-process}, which can be divided into five steps:
\begin{enumerate}[leftmargin=20pt,parsep=5pt,itemsep=0pt,topsep=5pt]
    \item \textbf{Cross-model word embedding alignment}: transfer the word embeddings of Liv4ever-MT to M2M100, enabling it to support Livonian.
    \item \label{item:step2} \textbf{4-lingual M2M training}: many-to-many translation training for all language pairs in $\{$En, Liv, Et, Lv$\}$, using only parallel data.
    \item \textbf{Synthetic data generation}: generate synthetic bi-text for En-Liv, using Et and Lv as pivot languages.
    \item \textbf{Combine data and retrain}: combine all the authentic and synthetic bi-text and retrain the model following step \ref{item:step2}.
    \item \textbf{Fine-tune \& post-process}: fine-tune the model on En$\Leftrightarrow$Liv using the validation set and perform online back-translation using monolingual data. Finally, apply rule-based post-processing to the model output.
\end{enumerate}

\subsection{Cross-model Word Embedding Alignment}
M2M100 1.2B does not support Livonian.
Therefore, we used Liv4ever-MT's SentencePiece model and vocabulary to process all the data.
For M2M100, the embeddings of new coming words can be randomly initialized.
However, randomly initialized word embeddings and the pre-trained models may not be compatible.
Inspired by supervised cross-lingual word embedding alignment~\cite{lample2018word}, we propose cross-model word embedding alignment (CMEA) to transform the trained word embeddings of Liv4ever-MT into M2M100, avoiding random initialization.

\paragraph{CMEA}
We denote Liv4ever-MT and M2M100 model by $l$ and $m$.
Their corresponding vocabularies and embedding matrices are $d_l, d_m$ and $\mathbf{X}^l, \mathbf{X}^m$.
\tablename~\ref{tab:vocab-stat} shows the statistics of the vocabularies.
\begin{table}[htpb]  
    \centering
    \begin{tabular}{c c c c}
        \toprule
            $|d_l|$ & $|d_m|$ & $|d_l \cap d_m|$ & $|d_l \cap d_m|/|d_l|$ \\
        \midrule
            47972  & 128108   & 11410 & 23.8\% \\
        \bottomrule
    \end{tabular}
    \caption{Statistics of Liv4ever-MT ($d_l$) and M2M100 ($d_m$) vocabularies.}
    \label{tab:vocab-stat}
\end{table}
Let $\mathbf{X}^f$ be the final embedding matrix we expected. We adopt $d_l$ as the final vocabulary, which can be divided into two parts: 
\begin{equation}
    d_l = (d_l\cap d_m) \cup (d_l-d_m).
\end{equation}
For the overlapped part $d_l\cap d_m$, $\mathbf{X}^f$ can reuse the embedding from $\mathbf{X}^m$:
\begin{equation}
    \mathbf{X}^f_{d_l\cap d_m} = \mathbf{X}^m_{d_l\cap d_m}.
\end{equation}
For the rest part $d_l-d_m$, we first find a liner transformation $\mathbf{W}$ between two embedding spaces such that:
\begin{equation}
\begin{aligned}
    \mathbf{W}^*=\argmin_{\mathbf{W}}&{\|\mathbf{W}\mathbf{X}^l_{d_l\cap d_m}-\mathbf{X}^m_{d_l\cap d_m}\|_F}
    \\
    &\myst  \mathbf{W}^T\mathbf{W}= \mathbf{I}.
\end{aligned}
\end{equation}
According to \citet{everson1998orthogonal},
\begin{equation}
\begin{aligned}
        \mathbf{W}^*&=\mathbf{UV}^T, \\
        \text{with } \mathbf{U} \mathbf{\Sigma} \mathbf{V}^{T}&=\operatorname{SVD}\left(\mathbf{X}^m_{d_l\cap d_m} {\mathbf{X}^l_{d_l\cap d_m}}^T\right).
\end{aligned}
\end{equation}
Then the word embeddings can be initialized as:
\begin{equation}   
    \mathbf{X}^f_{d_l- d_m} = \mathbf{W}^*\mathbf{X}^l_{d_l- d_m}.
\end{equation}

\paragraph{Experiment}
To investigate the effect of CMEA, we conducted \textbf{4-lingual M2M training} with different sampling temperature~\cite{aharoni-etal-2019-massively,tang-etal-2021-multilingual}.
\tablename~\ref{tab:4-lingual-m2m-training} shows the BLEU scores on the validation set.
We have the following observations:
\begin{itemize}[leftmargin=20pt,parsep=5pt,itemsep=0pt,topsep=5pt]
    \item M2M04 outperforms Liv4ever-MT by a large margin owing to the larger model size, more training data and the pre-trained parameters.
    
    \item On most language pairs, our proposed CMEA initialization significantly improves translation performance compared to random initialization of new coming embeddings.
    
    \item Temperature set to 5 with CMEA initialization achieves the best overall results.
    Therefore, we used this model in \textbf{synthetic data generation}.
\end{itemize}

\begin{table}[htpb]  
    \centering
    \setlength\tabcolsep{2.3pt}
    \begin{tabular}{l c c c c c c }
        \toprule
         & \multicolumn{2}{c}{\bf En-Liv} & \multicolumn{2}{c}{\bf Et-Liv} & \multicolumn{2}{c}{\bf Lv-Liv} \\
        \cmidrule(lr){2-3} \cmidrule(lr){4-5} \cmidrule(lr){6-7} 
        & $\Rightarrow$ & $\Leftarrow$ & $\Rightarrow$ & $\Leftarrow$ & $\Rightarrow$ & $\Leftarrow$ \\
        \midrule
        \makecell[l]{\bf Liv4ever-MT\\\citeauthor{rikters-etal-2022-machine}} &  14.3 & 19.3 & 20.5 & 24.4 & 22.3 & 29.3\\
        \midrule
        \bf M2M04 (T=5)         & 21.1 & 27.7 & 25.3 & 29.2 & 26.8 & 36.6  \\
        \bf ~~~~~~~~+ CMEA    & \bf23.0 & \bf28.4 & \bf27.2 & \bf30.7 & \bf28.5 & \bf37.6     \\
        \hdashline
        \bf M2M04 (T=10)        & \bf21.3 & 26.6 & 25.5 & 27.7 & 26.3 & 34.6 \\
        \bf ~~~~~~~~+ CMEA    & 21.1 & \bf27.1 & \bf26.0 & \bf29.6 & \bf27.5 & \bf36.3 \\
        \hdashline
        \bf M2M04 (T=20)        & 21.9 & 26.7 & \bf26.5 & \bf29.8 & 27.3 & \bf36.5 \\
        \bf ~~~~~~~~+ CMEA    & \bf22.1 & \bf27.4 & 25.8 & 27.9 & \bf27.9 & 33.8 \\
        \bottomrule
    \end{tabular}
    \caption{Experimental results of 4-lingual M2M training. We denote M2M04 as the 4-lingual translation model. `T' represents the sampling temperature.}
    \label{tab:4-lingual-m2m-training}
\end{table}

\subsection{Synthetic Data Generation}
Data agumentation~\cite{sennrich2016improving,jiao2020data,jiao2022exploiting,jiao2021self,he2022bridging}  is a widely used technique to boost the performance of neural machine translation. To augment the parallel data for En-Liv, we adopt both forward and backward translation to generate synthetic bi-text for En-Liv. 
\figurename~\ref{fig:training-process} (below) illustrates the process of synthetic data generation.

Considering the performances of Et/Lv$\Rightarrow$Liv are much better than En$\Rightarrow$Liv (see \tablename~\ref{tab:4-lingual-m2m-training}), we use Et and Lv as pivot languages to generate Liv instead of directly generating from En.
Taking Et as the pivot language, given authentic En-Et bi-text, we use the best model in \tablename~\ref{tab:4-lingual-m2m-training} to translate the Et into Liv, thus forming the synthetic En-Liv which is En-original.
Conversely, given authentic Et-Liv, we translate Et into En using Google Translate, forming the synthetic En-Liv which is Liv-original.
For Lv as the pivot language, we repeat the same steps.
\tablename~\ref{tab:statistics-of-synthetic-data} lists statistics of the synthetic En-Liv data after filtering.
\begin{table}[htpb]
    \centering
    \begin{tabular}{r r r}
    \toprule
    \multirow{2}{*}{\makecell{\bf Data Type}}     &  \multicolumn{2}{c}{\bf Pivot Language}\\
    \cmidrule{2-3}
         &  \multicolumn{1}{c}{Et} & \multicolumn{1}{c}{Lv} \\
    \midrule
    En-original   &  20.5M  & 11.2M\\
    Liv-original  &  14.2K  & 11.6K\\
    \bottomrule
    \end{tabular}
    \caption{The number of sentences of generated synthetic data after filtering, which is divided into four categories based on the original language and the pivot language.}
    \label{tab:statistics-of-synthetic-data}
\end{table}

\paragraph{Experiment}
We combine the authentic and synthetic bi-text and retrain the 4-lingual model.
The sampling temperature is set to 0 here to avoid downsampling for En-Liv.
When using only En-original or Liv-original synthetic data, we control the sampling frequency of the different language pairs to be consistent with using the full data.
\tablename~\ref{tab:retrain} shows the BLEU scores on the multi-way validation set.
We also report the round-trip BLEU on the monolingual En from the source of WMT22 En-De test set, which is En-original.
Unexpectedly, original-language greatly affects the model performance and causes inconsistent results between different evaluation methods:
\begin{itemize}[leftmargin=20pt,parsep=5pt,itemsep=0pt,topsep=5pt]
    \item En-original synthetic data remarkably degrades model performance on the validation set but significantly increases the round-trip BLEU.
    \item Liv-original synthetic data slightly reduces the performance on the validation set but moderately increases the round-trip BLEU.
    \item When using both kinds of data, the best round-trip BLEU is achieved.
    However, the performance on the validation set is still worse than the baseline.
\end{itemize}

\begin{table}[htpb]  
    \centering
    \setlength\tabcolsep{1.5pt}
    \begin{tabular}{l c c  c}
        \toprule
        & \multicolumn{2}{c}{\bf Valid (multi-way)} & \multirow{2}{*}{\makecell{\bf Round-Trip\\ \bf(En-original)}}\\
        \cmidrule(lr){2-3}
        & \bf En$\Rightarrow$Liv & \bf Liv$\Rightarrow$En  \\
        \midrule
        \makecell{\bf M2M04 (T=5)\\\bf+CMEA} & 23.0 & 28.4 & 23.4 \\
        \midrule
        \multicolumn{4}{c}{\bf Add synthetic data and retrain} \\
        \bf En-original  & 17.2 & 17.5 & 30.7 \\
        \bf Liv-original & 21.5 & 27.4 & 25.8 \\
        \bf Both         & 17.0 & 19.3 & 32.7 \\
        
        \bottomrule
    \end{tabular}
    \caption{Translation performance after adding the synthetic data and retraining the model.}
    \label{tab:retrain}
\end{table}
As described in Section~\ref{sec:round-trip-bleu}, we consider round-trip BLEU the more appropriate evaluation in this competition due to more data, more general domain, and the same original language as the WMT22 En-Liv test set.
Therefore, we used both kinds of synthetic data in our submissions. 

\subsection{Fine-tuning \& Post-processing}
\paragraph{Fine-tuning}
To further exploit the bilingual and monolingual data, we fine-tuned the model on the En$\Leftrightarrow$Liv validation set for 500 steps jointly with online back-translation on monolingual data.
\paragraph{Post-processing}
We apply the following rule-based post-processing:
\begin{itemize}[leftmargin=10pt,parsep=5pt,itemsep=0pt,topsep=5pt]
    \item Apply NFC normalization.
    \item Replace all the \texttt{httpshttp} with \texttt{https://}.
    \item Replace \texttt{<unk>} with empty string.
    \item When a comma appears between two digits, replace it with a decimal point (only for Liv).
    \item Regenerate the sentences that detected as repetition with no-repeat constraint\footnote{We use \texttt{----no-repeat-ngram-size 2} in fairseq-generate.} (only for Liv).
\end{itemize}

\paragraph{Final results}
\tablename~\ref{tab:fine-tune} shows the test set performance and round-trip BLEU after fine-tuning and post-processing.
As seen, fine-tuning significantly improves model performance on both test set and round-trip BLEU.
Post-processing further boosts the performance on the test set.

\begin{table}[htpb]
    \centering
    \begin{tabular}{l c c c}
        \toprule
        & \multicolumn{2}{c}{\bf Test Set} & \multirow{3}{*}{\makecell{\bf Round-Trip\\\bf BLEU}}\\
        & \multicolumn{2}{c}{\bf En-Liv}   \\
        \cmidrule(lr){2-3}
        & \bf $\Rightarrow$ & \bf $\Leftarrow$  \\
        \midrule
        \makecell[l]{\bf Before\\\bf fine-tuning} & 15.8 & 29.4 & 32.7 \\
        \hdashline
        \bf +Fine-tuning     & 16.3 & 30.1 & 37.1 \\
        \bf +Post-proc.      & 17.0 & 30.4 & 37.1 \\

        \bottomrule
    \end{tabular}
    \caption{Translation performance after fine-tuning and post-processing.}
    \label{tab:fine-tune}
\end{table}

\section{Conclusion}
\label{sec:conclusion}
This paper presents the Tencent AI Lab - Shanghai Jiao Tong University (TAL-SJTU) Low-Resource Translation systems for the WMT22 shared task.
We start from the M2M100 1.2B model and investigate techniques to adapt it to English$\Leftrightarrow$Livonian.
We propose cross-model word embedding alignment that transfer the embeddings of Liv4ever-MT to M2M100, enabling it to support Livonian.
Then, Estonian and Latvian are involved in model training and synthetic data generation as auxiliary and pivot languages.
We further fine-tune the model with validation set and online back-translation followed by rule-based post-processing.
In model evaluation, we correct the inaccurate evaluation of Livonian due to inconsistent Unicode normalization and use round-trip BLEU as an alternative to the standard validation set.

\section*{Acknowledgements}
Zhiwei He and Rui Wang are with MT-Lab, Department of Computer Science and Engineering, School of Electronic Information and Electrical Engineering, and also with the MoE Key Lab
of Artificial Intelligence, AI Institute, Shanghai Jiao Tong University, Shanghai 200204, China.
Rui is supported by General Program of National Natural Science Foundation of China (6217020129), Shanghai Pujiang Program (21PJ1406800), and Shanghai Municipal Science and Technology Major Project (2021SHZDZX0102).
Zhiwei is supported by CCF-Tencent Open Fund (RAGR20210119).

\bibliography{custom}
\bibliographystyle{acl_natbib}

\newpage
\appendix
\section{Re-evaluating Liv4ever-MT}
\tablename~\ref{tab:re-evluating-liv4ever-mt} shows the results of re-evaluating Liv4ever-MT on all language pairs.
Normalizing references to NFKC improves the average BLEU scores by +2.54 on the entire set and +8.26 on the Satversme subset.
It is worth mentioning that liv4ever-test contains data from the following sources: Facebook, Livones.net, Dictionary, Trilium, Stalte, JEFUL and Satversme.
However, there does not exist the Unicode inconsistency problem in the other sources except Satversme.

\label{sec-appendix:re-evluating-liv4ever-mt}
\begin{table*}[!htb]
    \centering
    \setlength\tabcolsep{2.6pt}
    \begin{tabular}{l | c c c c c c c c c c c c c}
        \toprule
         & \multicolumn{3}{c}{\bf XX$\Rightarrow$En} & \multicolumn{3}{c}{\bf XX$\Rightarrow$Et} & \multicolumn{3}{c}{\bf XX$\Rightarrow$Lv} & \multicolumn{3}{c}{\bf XX$\Rightarrow$Liv} & \multirow{2}{*}{\bf Avg.}\\
        \cmidrule(lr){2-4} \cmidrule(lr){5-7} \cmidrule(lr){8-10} \cmidrule(lr){11-13} 
        \makecell[c]{\bf XX} & \bf Et & \bf Lv & \bf Liv & \bf En & \bf Lv & \bf Liv & \bf En & \bf Et & \bf Liv & \bf En & \bf Et & \bf Lv \\
        \midrule
        & \multicolumn{12}{c}{\bf All} \\
        \makecell[l]{\bf Liv4ever-MT\\(\citeauthor{rikters-etal-2022-machine})} &  26.17 & 21.53 & 19.01 & 19.48 & 22.38 & 23.05 & 20.85 & 23.44 & 25.24 & 11.03 & 16.40 & 17.65 & 20.52\\
        \hdashline
        \bf Our Eval.   & 25.90 & 17.94 & 18.90 & 19.28 & 22.31 & 22.86 & 20.20 & 23.31 & 24.88 & 10.90 & 16.62 & 17.69 & 20.07\\
        \bf + Norm Ref. & \bf 26.20 & \bf 18.06 & \bf 19.26 & \bf 20.72 & \bf 24.28 & \bf 24.42 & \bf 24.10 & \bf 27.77 & \bf 29.33 & \bf 14.31 & \bf 20.51 & \bf 22.35 & \bf 22.61\\

        \midrule
        & \multicolumn{12}{c}{\bf Subset (Satversme)} \\
        \makecell[l]{\bf Liv4ever-MT\\(\citeauthor{rikters-etal-2022-machine})} & - & - & 24.49 & - & - & - & - & - & - & 7.69 & - & - & -\\
        \hdashline
        \bf Our Eval.   & 27.50 & 19.77 & 24.68 & 16.69 & 20.22 & 18.68 & 16.05 & 15.10 & 19.38 & 7.58 & 7.18 & 9.23 & 16.83\\

        \bf + Norm Ref. & \bf 28.45 & \bf 20.21 & \bf 25.76 & \bf 21.41 & \bf 26.74 & \bf 23.75 & \bf 29.10 & \bf 29.82 & \bf 33.56 & \bf 18.23 & \bf 19.87 & \bf 24.15 & \bf 25.09\\
        \bottomrule
    \end{tabular}
    \caption{BLEU scores of Liv4ever-MT on liv4ever-test. \textbf{Liv4ever-MT}~(\citeauthor{rikters-etal-2022-machine}): copied from~\citet{rikters-etal-2022-machine}. \textbf{Our Eval.}: We use the released Liv4ever-MT to generate translation outputs and re-evaluate them with the original references, which shows similar results compared with \citet{rikters-etal-2022-machine}.
    \textbf{+ Norm. Ref.}: re-evaluation after normalizing the encoding of references to NFKC.}
    \label{tab:re-evluating-liv4ever-mt}
\end{table*}
\end{document}